\documentclass[runningheads]{llncs}
\usepackage[T1]{fontenc}
\usepackage{graphicx}
\usepackage{amssymb}
\usepackage{amsmath}
\usepackage{hyperref}
\usepackage{multirow}
\usepackage{pdflscape}
\usepackage{subfig}
\usepackage{lipsum}
\usepackage{graphicx}
\usepackage{adjustbox}
\usepackage[utf8]{inputenc}
\usepackage[normalem]{ulem}
\usepackage{fontspec}
\usepackage{polyglossia}
\usepackage{amsmath}
\usepackage{bbm}
\usepackage{algorithm}
\usepackage{booktabs} 
\usepackage{array}
\usepackage{xcolor}
\usepackage{caption}
\usepackage{algpseudocode}%
\usepackage{listings}%
\usepackage{subcaption}
\usepackage{algpseudocode}
\usepackage[ruled,vlined,algo2e]{algorithm2e}
\usepackage{diagbox}
\begin{document}
%
%
%
\title{Evaluating Cross-Lingual Classification approaches Enabling Topic Discovery for Multilingual Social Media Data}


\newcommand{\anonymize}[1]{}

\author{
Deepak Uniyal\inst{1}\orcidID{0000-0002-7665-1585} \and
Md Abul Bashar\inst{2}\orcidID{0000-0003-1004-4085} \and
Richi Nayak\inst{3}\orcidID{0000-0002-9954-0159}
}

%

\institute{
Centre for Data Science, School of Computer Science,\\
Queensland University of Technology, Brisbane, QLD 4000, Australia \\[1.2ex]
$^{1}$\texttt{deepak.uniyal@hdr.qut.edu.au}, 
$^{2}$\texttt{m1.bashar@qut.edu.au}, 
$^{3}$\texttt{r.nayak@qut.edu.au}
}
\maketitle              

\begin{abstract}
Analysing multilingual social media discourse remains a major challenge in natural language processing, particularly when large-scale public debates span across diverse languages. This study investigates how different approaches for cross-lingual text classification can support reliable analysis of global conversations. Using hydrogen energy as a case study, we analyse a decade-long dataset of over nine million tweets in English, Japanese, Hindi, and Korean (2013--2022) for topic discovery. The online keyword-driven data collection results in a significant amount of irrelevant content. We explore four approaches to filter relevant content: (1) translating English annotated data into target languages for building language-specific models for each target language, (2) translating unlabelled data appearing from all languages into English for creating a single model based on English annotations, (3) applying English fine-tuned multilingual transformers directly to each target language data, and (4) a hybrid strategy that combines translated annotations with multilingual training. Each approach is evaluated for its ability to filter hydrogen-related tweets from noisy keyword-based collections. Subsequently, topic modeling is performed to extract dominant themes within the relevant subsets. The results highlight key trade-offs between translation and multilingual approaches, offering actionable insights into optimising cross-lingual pipelines for large-scale social media analysis.

\end{abstract}

\section{Introduction}

The growth of social media platforms such as Twitter\footnote{Recently changed its name to $\mathbb{X}$. In this paper, we refer to this platform as Twitter.} has enabled the large-scale collection of user-generated content across multiple languages, offering opportunities for studying global public opinion \cite{uniyal2024twitter}. A key challenge in this domain is relevance classification, i.e. identifying tweets pertinent to a specific topic across diverse linguistic contexts. The multilingual nature of such data introduces significant challenges for natural language processing (NLP), including handling imbalances in language resources, preserving cultural nuances, and mitigating translation-induced noise \cite{xu2022survey}. These issues fuel an ongoing debate: Should multilingual social media data be processed using separate native language-specific models, translated into a pivot language (e.g., English), or directly handled by multilingual models such as multilingual BERT (mBERT) \cite{devlin2018bert} and XLM-RoBERTa (XLM-R) \cite{ruder2019unsupervised} \cite{liu2024translation} \cite{isbister2021should}.

While NLP research has advanced multilingual text processing methods, the optimal strategy remains context-dependent. Translating all data to a pivot language like English \cite{kumar2023zero} simplifies downstream tasks but risks losing language-specific subtleties. Training separate language-specific models can preserve these nuances, yet it demands extensive annotated data for each language, which is often unavailable. Multilingual transformer models, such as mBERT \cite{devlin2018bert} and XLM-R \cite{ruder2019unsupervised}, offer the ability to process multiple languages directly; however, their performance varies depending on the training resources available for specific languages, often declining for low-resource language pairs \cite{pires2019multilingual}.

To address these challenges, this study systematically evaluates four cross-lingual relevance classification approaches for large-scale, multi-year Twitter data. The dataset for this study comprises a large corpus of Twitter data spanning multiple years, collected in English, Japanese, Korean, and Hindi. However, a critical constraint is the limited availability of annotated data, a common challenge in real-world scenarios due to resource limitations. Only a small set of 5,000 English tweets is annotated for relevance, while the vast majority of the data, comprising millions of unlabelled tweets across all four languages remains unannotated. This scarcity of labeled data complicates the development of robust cross-lingual systems, necessitating innovative approaches to leverage the available annotations effectively.

The first approach employs language-specific modeling, translating the 5,000 English-annotated tweets into Japanese, Korean, and Hindi to train separate classifiers, while using the original English annotations for an English model. The second approach adopts a translation-based pipeline, translating all non-English tweets into English and applying a single English classifier. The third approach leverages mBERT, fine-tuned on the English annotations and applied in a zero-shot manner \cite{yin2019benchmarking} to original-language tweets. Finally, the hybrid approach combines translation and multilingual training, translating English annotations into multiple languages, merging them into a single multilingual training set, and fine-tuning mBERT for cross-lingual prediction.

The main contributions of this study are as follows:
\begin{enumerate}
    \item Cross-Lingual Classification. A comparative evaluation of four approaches for relevance classification in large-scale multilingual Twitter data, enhancing the filtering of hydrogen-energy content.
    \item Key Themes and Phrases. An integrated pipeline using topic modeling to uncover thematic trends over a decade.
    \item Temporal Evolution. Insights into the evolution of hydrogen energy discourse from 2013 to 2022, including periods of heightened user engagement.
    \item First Decade-long Multilingual Analysis. A pioneering decade-long, multilingual analysis of hydrogen energy discourse across English, Japanese, Hindi, and Korean, with a scalable framework for future clean energy studies.
\end{enumerate}

The paper is organised as follows: Section~\ref{LiteratureReview} covers related work, Section~\ref{methodology} details our methodology, Section~\ref{resultsAndDiscussion} presents results, and Section~\ref{conclusion} offers conclusions and future directions.

\section{Literature Review}
\label{LiteratureReview}

Cross-lingual NLP has progressed from early translation-based methods to advanced multilingual models, addressing the challenge of linguistic diversity in multilingual datasets. The foundational ``translate-then-classify" approach, pioneered by early work in the field, translates source text into a pivot language (e.g., English) to enable processing, though its effectiveness hinges on translation accuracy and struggles with semantic nuances \cite{wan2009co}. Subsequent advancements introduced cross-lingual alignment techniques, such as structural correspondence learning, which identifies consistent pivot features across languages, reducing reliance on machine translation quality \cite{prettenhofer2011cross}. The development of multilingual word embeddings further enhanced semantic similarity across languages through linear transformations \cite{mikolov2013efficient}.
The advent of transformer-based models marked a significant leap, with Multilingual BERT (mBERT) demonstrating robust cross-lingual performance in tasks like classification and entity recognition \cite{pires2019multilingual}. Enhanced models like XLM and XLM-R, trained on larger corpora with refined objectives, have improved transfer capabilities across diverse languages \cite{conneau2019cross} \cite{conneau2019unsupervised}. However, these models often underperform with low-resource languages. Recent hybrid approaches combine monolingual specialisation, which captures language-specific nuances, with multilingual generalisation for cross-lingual alignment, though their application in specialised domains remains underexplored.

Our study addresses this gap by evaluating four cross-lingual relevance classification approaches -- monolingual, translation-based, multilingual, and hybrid -- on a decade-long, four-language dataset, enabling effective filtering for topic modeling to uncover regional hydrogen energy themes.

\begin{figure}[ht]
\centering
\includegraphics[width=0.95\textwidth]{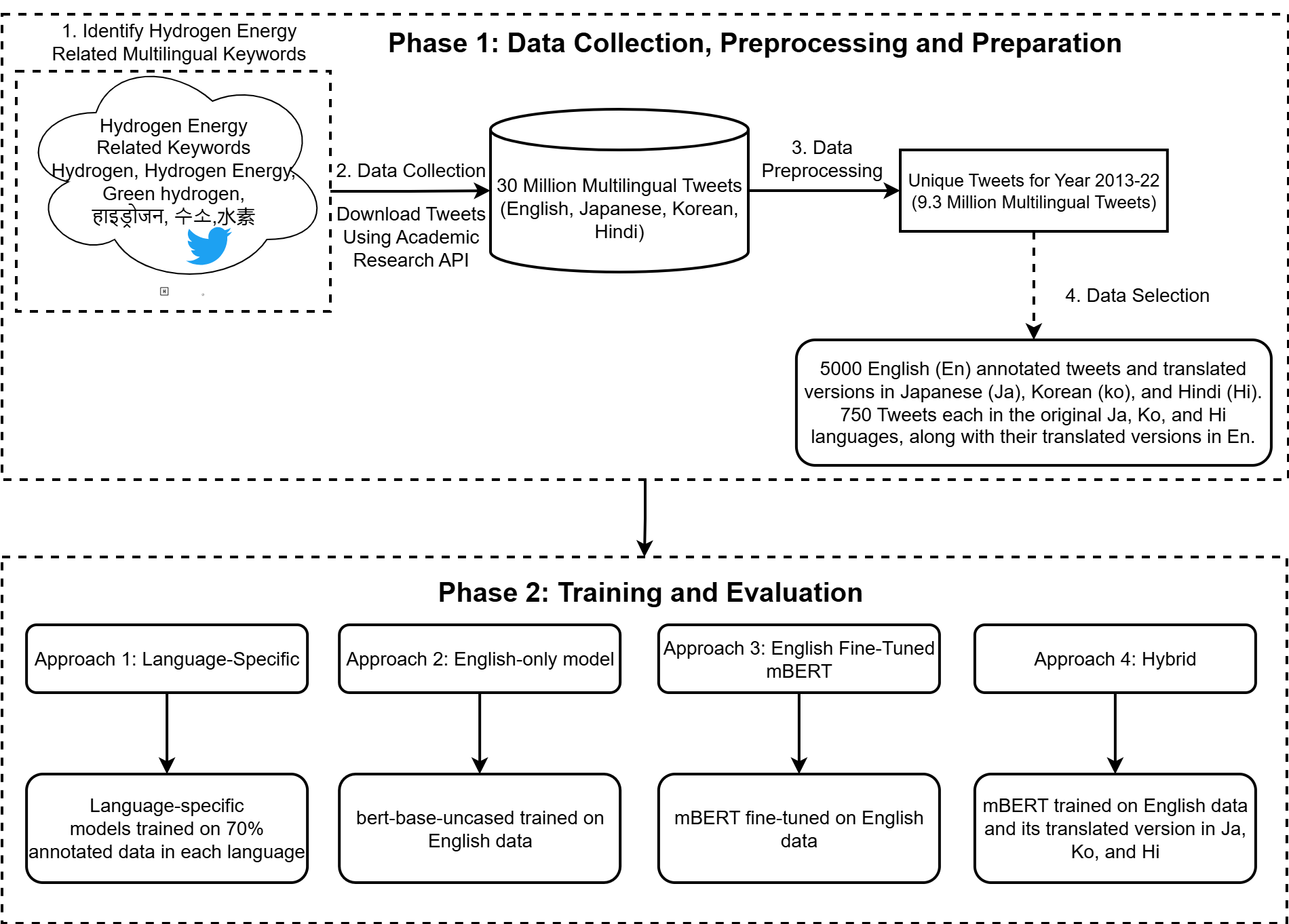}
\caption{Architecture of the proposed multi-step methodology, including data collection, data preprocessing, and relevance classification.} 
\label{fig:methodology}
\end{figure}

\section{Methodology}
\label{methodology}
This study presents a systematic pipeline for analysing multilingual Twitter discourse on hydrogen energy spanning 2013 to 2022. The methodology, illustrated in Figure \ref{fig:methodology}, integrates data collection, preprocessing, and relevance classification, which enable further NLP tasks such as topic modeling or sentiment mining, based on the best method. The central challenge arises from the multilingual nature of the corpus: Tweets appear in English (EN), Japanese (JA), Korean (KO), and Hindi (HI), while supervised annotations are initially available only in English.

To address this challenge, we developed and evaluated four distinct cross-lingual relevance classification approaches: (1) Training independent language-specific models for each language, (2) Training a single English model on English annotated data, (3) Fine-tuning mBERT on English annotated data and employing zero-shot predictions for other languages, and (4) Fine-tuning mBERT on English annotated data and its translation into other languages, followed by cross-lingual predictions.

Following relevance classification using the optimal approach, topic modelling is performed on the filtered, thematically coherent dataset.

\subsection{Data Collection and Preprocessing}
\label{dataCollection}
We collected a large-scale dataset of 30.7 million tweets related to hydrogen energy using the Academic Twitter API v2, covering the period from January 2013 to December 2022. Keywords were curated through a combination of literature review and expert consultation to ensure comprehensive coverage of the discourse across four languages: English, Japanese, Hindi, and Korean \cite{uniyal2024twitter}.

The initial processing, as described in \cite{uniyal2024twitter}, included steps such as metadata mapping, deduplication, language-specific text normalisation, and tokenisation. After removing duplicate tweets, we retained approximately 9.9 million unique tweets. To ensure temporal consistency, we further filtered tweets strictly posted between 2013 and 2022. This step excluded retweets and other content originating outside the target window but was re-shared during the period. The resulting dataset consisted of approximately 9.3 million tweets.

For supervised relevance classification, a labeled dataset of 5{,}000 English tweets was created, uniformly sampled across years to maintain temporal coverage. Each tweet was annotated as \textit{relevant} or \textit{irrelevant} by three domain experts following detailed guidelines. Inter-annotator reliability, measured using Fleiss’ kappa \cite{fleiss1971measuring}, achieved a score of 0.85, indicating strong agreement. Approximately 45\% of tweets were labeled as irrelevant, highlighting the necessity of automated filtering. This annotated dataset provides the foundation for training all subsequent relevance classification approaches. 

As we lack a labeled dataset in other languages, we translated these 5{,}000 tweets into Japanese, Korean, and Hindi using \texttt{facebook/\allowbreak nllb-200-\allowbreak distilled-\allowbreak 600M} \cite{costa2022no} neural machine translation model. For a robust evaluation, we randomly selected and manually annotated 750 tweets each from original, non-translated Japanese, Hindi, and Korean tweets, ensuring uniform distribution across years and referencing their translation for accuracy.

\begin{table}[t]
\centering
\scriptsize
\caption{Training and testing setup for four relevance-classification approaches.}
\renewcommand{\arraystretch}{1.2}
\setlength{\tabcolsep}{5pt}
\begin{tabular}{@{}p{2.2cm}p{3.0cm}p{2.3cm}p{3.6cm}@{}}
\toprule
\textbf{Approach} & \textbf{Training Data} & \textbf{Test Data} & \textbf{Model Setup} \\ 
\midrule
\textbf{1. Language-specific} & 5{,}000 English annotated tweets and their translations (JA, KO, and HI)\textsuperscript{a} & \multirow{4}{2.6cm}{750 tweets each language taken from original language data} & Separate language-specific models for each language\\

\textbf{2. English-only} &  5{,}000 English annotated tweets &  & Single English model; all tweets translated to English \\

\textbf{3. English Fine-Tuned mBERT} &   5{,}000 English annotated tweets & & Single mBERT model fine-tuned on English data; zero-shot prediction on other non-English languages \\

\textbf{4. Hybrid} & 5{,}000 English annotated tweets and their translations (JA, KO, and HI)\textsuperscript{a} & & Single mBERT model trained on combined multilingual data \\

\bottomrule
\end{tabular}
\vspace{-0.1em}
\begin{flushleft}
\footnotesize
\textsuperscript{a} JA = Japanese, KO = Korean, HI = Hindi.
\end{flushleft}
\label{tab:four_approaches}
\end{table}

\subsection{Multilingual Relevance Classification}
The primary challenge in classifying the relevance of this dataset stems from its multilingual composition, with initial labels available only for the English subset. To address this, we developed and evaluated four distinct cross-lingual relevance classification approaches, which vary based on the training data composition and the type of encoder used (monolingual or multilingual), as detailed in Table~\ref{tab:four_approaches}. We first outline the experimental setup and data utilization across all approaches, followed by a detailed description of each approach:

\subsubsection{Experimental setup} Our experimental framework incorporates a ten-run with ten different seeds, where the data for each language is divided into a train, validation and test sets in a ratio of 70\%, 15\%, and 15\%, respectively. Depending on the chosen approach, a suitable subset of data is selected to train the model across ten seeds, and the best model is saved for prediction. We implement early stopping with a patience of 5 epochs, terminating training if no improvement in validation accuracy is observed.

We evaluate the performance using accuracy and F1-score, as these metrics are widely used in text classification \cite{jointmatch}. Both metrics are reported as mean and standard deviation over ten runs to ensure a robust evaluation.

Since the original annotations were available only for English data, we created a uniform set of 750 annotated tweets per language (English, Japanese, Korean, and Hindi) from original language tweets, used exclusively for testing to evaluate model performance on authentic unseen tweets across all languages. These annotated samples are used to rigorously test each approach and report comparative performance metrics.

\subsubsection{Approach 1: Monolingual Models} \label{approach1}
In the first approach, we use 5{,}000 English annotated tweets and their translations for all languages to fine-tune separate monolingual classifiers independently. Each classifier is trained on its respective language data using a language-specific transformer model: 
\begin{enumerate}
    \item English: \texttt{bert-\allowbreak base-\allowbreak uncased} \cite{devlin2018bert}, a 12-layer bidirectional Transformer encoder with 110M parameters, pre-trained on BooksCorpus and English\\ Wikipedia.
    \item Japanese:  \texttt{cl-\allowbreak tohoku/ \allowbreak bert-\allowbreak base-\allowbreak japanese} \cite{bert_japanese_repo}, a BERT-base model pre-trained on Japanese Wikipedia, tokenised with a WordPiece-based vocabulary optimised for Japanese morphology.
    \item Korean: \texttt{monologg/kobert} \cite{park2019distilkobert}, a Korean BERT variant trained on a large-scale Korean corpus, adapted with a custom tokeniser to handle Korean morphology and spacing.
    \item Hindi: \texttt{ai4bharat/indic-bert} \cite{kakwani2020indicnlpsuite}, a multilingual BERT model covering 12 major Indian languages, trained on IndicCorp and designed for resource-limited Indic NLP tasks.
\end{enumerate}

For each model, the architecture consists of the language-specific encoder followed by a dropout layer ($p=0.3$), a fully connected projection layer, a \texttt{tanh} activation, and a softmax output layer. Fine-tuning is performed independently for each language on corresponding labeled data.
The performance of each model is tested on the uniform set of 750 annotated tweets per language from original language data.

\subsubsection{Approach 2: English-only Model} \label{approach2}
In the second approach, we utilise annotated English tweets to train the English-only model, i.e. \texttt{bert-\allowbreak base-\allowbreak uncased}. The model architecture mirrors that of approach~1 (\ref{approach1}) for prediction purposes. The test is performed on the English portion of 750 annotated tweets, and additional tests are conducted on the English-translated versions of the 750 annotated tweets per language (Japanese, Korean, and Hindi) from the original language data.

\subsubsection{Approach 3: Multilingual Model} \label{approach3}
In the third approach, we adopt a fully multilingual strategy by leveraging the \texttt{bert-\allowbreak base-\allowbreak multilingual-\allowbreak uncased}\\model, referred as mBERT \cite{sanchez2023cross}, which supports 104 languages including English, Japanese, Korean, and Hindi. The fine-tuning phase utilises English annotated tweets to fine-tune the mBERT, relying on its generalisation capabilities from pre-training. The model architecture consists of the pre-trained mBERT encoder followed by a dropout layer with $p=0.3$, a fully connected projection layer, a \texttt{tanh} activation function, and a final softmax classification layer producing binary relevance predictions on the input data. The test is performed on the English portion of 750 annotated tweets per language, using the best model of ten runs for prediction. Additional tests are performed for non-English languages (Japanese, Korean, and Hindi) using the test set of 750 tweets in their native form. 

\subsubsection{Approach 4: Hybrid Multilingual Model} \label{approach4} The fourth approach combines the benefits of the multilingual approach as in approach~3 (\ref{approach3}), which relies on mBERT fine-tuned on English data with non-English translated data enriched with annotations, integrating these elements into a unified training framework.
These translated annotations, together with the original English annotated data form a combined multilingual training set. This dataset is used to fine-tune an mBERT classifier with the same architecture as in approach~3. By using a multilingual dataset for fine-tuning, this approach aims to improve the classification performance on non-English tweets while preserving the advantages of mBERT’s cross-lingual representation learning. The test is performed on the uniform set of 750 annotated tweets per language from original language data, and the best model from ten runs is saved for inference. 

\subsection{Topic Modelling}
\label{topicModelling}
Experimental results show that Approach 2 (the model trained with English-annotated tweets only) provided the most accurate relevancy distinction. After identifying \texttt{approach~2} as the best-performing and most interpretable strategy for relevance classification, we used its filtered output for topic modelling. One key advantage of this approach is that all tweets, regardless of their original language, are translated into English during the classification stage. This not only enables consistent processing but also makes the resulting themes more accessible to a wider audience by presenting the discourse in a single language. Tweets labeled as \textit{irrelevant} were excluded from further analysis, ensuring that only content truly related to hydrogen energy shaped the thematic structure. We explored topics from other approaches in their original languages, but their quality was noisy, likely due to challenges in handling multilingual data effectively.

For topic discovery, we selected Non-negative Matrix Factorisation (NMF) \cite{balasubramaniam2020understanding} over methods like Latent Dirichlet Allocation (LDA) due to its interpretability, robustness, and superior alignment with human judgment \cite{egger2022topic}. NMF effectively derived coherent topics from sparse, high-dimensional TF-IDF representations, which incorporated unigrams, bigrams, and trigrams, filtered infrequent terms, and limited the feature space to 10{,}000 terms for computational efficiency. Configured to extract up to ten topics per language with reproducible initialisation, NMF ensured consistent comparability across datasets.

This pipeline produced topic keywords with associated weights, tweet-to-topic assignments identifying each tweet’s dominant theme, and temporal distributions tracking yearly frequencies from 2013 to 2022. These outputs revealed the evolution of hydrogen energy discourse, highlighting periods of thematic growth or decline. By relying on a unified English-translated dataset, the approach balanced model performance, interpretability, and comparability, making the analysis accessible to both technical and non-technical audiences while upholding rigorous topic extraction and trend analysis.

\section{Results and Discussion}
\label{resultsAndDiscussion}
This section provides a detailed analysis of the outcomes from the cross-lingual relevance classification and topic modeling applied to Twitter data on hydrogen energy discourse. The analysis focuses on temporal trends, including tweet distribution over years and topic evolution, utilising the processed dataset derived from \texttt{approach~2}. The findings are interpreted in the context of hydrogen energy market trends, policy developments, and technological advancements, informed by prior studies such as \cite{uniyal2024twitter}.

\subsection{Performance Analysis of Cross-Lingual Relevance Classification Approaches}

Table \ref{tab:approach_performance} presents test performance metrics (accuracy and F1-score) for four cross-lingual relevance classification approaches, evaluated on a uniform test set of 750 tweets per language (English, Japanese, Korean, and Hindi) from original language data. The results, expressed with standard deviations across five runs, highlight the strengths and limitations of each approach, offering insights into their suitability for multilingual tweet analysis.

\begin{table}[ht]
\scriptsize
\centering
\caption{Test accuracy and F1-score for four cross-lingual relevance classification methods. Best results are shown in bold blue; second-best in bold.}
\label{tab:approach_performance}
\begin{tabular}{llccc}
\toprule
\textbf{Approach} & \textbf{Language} & \textbf{Model} & \textbf{Accuracy (\%)} & \textbf{F1-score (\%)} \\
\midrule
\multirow{4}{*}{Approach 1: Language-Specific} 
 & EN & \texttt{BERT} & \textcolor{blue}{\textbf{97.72 ± 0.23}} & \textcolor{blue}{\textbf{97.70 ± 0.23}} \\
 & JA & \texttt{Japanese BERT} & 75.79 ± 0.49 & 74.88 ± 0.56 \\
 & KO & \texttt{Korean BERT} & \textcolor{blue}{\textbf{91.40 ± 0.31}} & \textcolor{blue}{\textbf{91.35 ± 0.31}} \\
 & HI & \texttt{Hindi BERT} & 78.08 ± 1.10 & 78.04 ± 1.10 \\
\midrule

\multirow{4}{*}{Approach 2: English-only} 
 & EN & \texttt{BERT} & \textcolor{blue}{\textbf{97.72 ± 0.23}} & \textcolor{blue}{\textbf{97.70 ± 0.23}} \\
 & JA & \texttt{BERT} &\textcolor{blue}{\textbf{79.85 ± 0.66}} & \textcolor{blue}{\textbf{79.30 ± 0.69}} \\
 & KO & \texttt{BERT} & \textbf{86.03 ± 0.22} & \textbf{85.78 ± 0.23} \\
 & HI & \texttt{BERT} & \textcolor{blue}{\textbf{90.59 ± 0.27}} & \textcolor{blue}{\textbf{90.53 ± 0.28}} \\
\midrule
\multirow{4}{*}{Approach 3: English Fine-Tuned mBERT} 
 & EN & \texttt{mBERT} & \textbf{96.68 ± 0.28} & \textbf{96.66 ± 0.28} \\
 & JA & \texttt{mBERT} & 50.21 ± 0.14 & 33.80 ± 0.30 \\
 & KO & \texttt{mBERT} & 50.56 ± 0.18 & 34.56 ± 0.39 \\
 & HI & \texttt{mBERT} & 53.59 ± 0.28 & 40.97 ± 0.55 \\
\midrule
\multirow{4}{*}{Approach 4: Hybrid} 
 & EN & \texttt{mBERT} & 94.81 ± 0.58 & 94.79 ± 0.58 \\
 & JA & \texttt{mBERT} & \textbf{78.49 ± 0.44} & \textbf{77.78 ± 0.49} \\
 & KO & \texttt{mBERT} & 83.57 ± 0.52 & 83.27 ± 0.54 \\
 & HI & \texttt{mBERT} & \textbf{81.76 ± 0.48} & \textbf{81.18 ± 0.52} \\
\bottomrule
\end{tabular}
\end{table}

\textbf{Approach 1 (Language-Specific)} leverages separate language-specific\\BERT models, trained on English-annotated tweets and their translations, with performance assessed on the test set. It achieves the highest accuracy and F1-score for English (97.72\% ± 0.23\% accuracy, 97.70\% ± 0.23\% F1-score), reflecting the robustness of the BERT model on its native training language. For non-English languages, performance varies: Japanese (75.79\% ± 0.49\% accuracy, 74.88\% ± 0.56\% F1-score), Korean (91.40\% ± 0.31\% accuracy, 91.35\% ± 0.31\% F1-score), and Hindi (78.08\% ± 1.10\% accuracy, 78.04\% ± 1.10\% F1-score) show that Korean benefits most from the language-specific model, while Japanese and Hindi exhibit lower and more variable performance, possibly due to translation errors or linguistic complexity.

\textbf{Approach 2 (English-only)} relies on a single BERT model trained on English-annotated tweets, applied to the English portion of the test set and translated versions of non-English tweets. It excels in English (97.72\% ± 0.23\% accuracy, 97.70\% ± 0.23\% F1-score) and demonstrates strong performance on non-English languages: Japanese (79.85\% ± 0.66\% accuracy, 79.30\% ± 0.69\% F1-score), Korean (86.03\% ± 0.22\% accuracy, 85.78\% ± 0.23\% F1-score), and Hindi (90.59\% ± 0.27\% accuracy, 90.53\% ± 0.28\% F1-score). This approach outperforms Approach 1 for Japanese and Hindi, suggesting that translation to English provides a more consistent feature representation, likely due to high-quality translation and the effective model available in English language.

\textbf{Approach 3 (English Fine-Tuned mBERT)} uses mBERT fine-tuned on English annotated tweets in a zero-shot setting. It performs well on English (96.68\% ± 0.28\% accuracy, 96.66\% ± 0.28\% F1-score) but shows significantly degraded performance on non-English languages: Japanese (50.21\% ± 0.14\% accuracy, 33.80\% ± 0.30\% F1-score), Korean (50.56\% ± 0.18\% accuracy, 34.56\% ± 0.39\% F1-score), and Hindi (53.59\% ± 0.28\% accuracy, 40.97\% ± 0.55\% F1-score). The low F1-scores (around 34-41\%) compared to accuracy (50-54\%) indicate a severe precision-recall imbalance, highlighting mBERT's inability to generalize from English to diverse linguistic structures without multilingual training.

\textbf{Approach 4 (Hybrid)} enhances mBERT with a combined training set of English annotated tweets and their translations, improving cross-lingual performance. It achieves balanced results: English (94.81\% ± 0.58\% accuracy, 94.79\% ± 0.58\% F1-score), Japanese (78.49\% ± 0.44\% accuracy, 77.78\% ± 0.49\% F1-score), Korean (83.57\% ± 0.52\% accuracy, 83.27\% ± 0.54\% F1-score), and Hindi (81.76\% ± 0.48\% accuracy, 81.18\% ± 0.52\% F1-score). This approach outperforms Approach 3 across all non-English languages and shows competitive performance with Approach 1 for non-English languages, effectively bridging monolingual and multilingual strategies.

In general, Approach 1 excels in English and Korean, but shows variability in Japanese and Hindi. Approach 2 provides the highest overall performance, particularly for English, Japanese, and Hindi, making it suitable for translation-based workflows. Approach 3 demonstrates the limitations of zero-shot multilingual transfer, while Approach 4 offers the balanced solution, leveraging multilingual training to improve consistency across languages. For practical deployment, Approach 2 is ideal for translation-heavy scenarios, whereas Approach 4 provides a robust foundation for truly multilingual applications, especially where native language data quality varies.

\begin{table}[t]
\centering
\scriptsize
\caption{Distribution of Relevant (R) and Irrelevant (I) Tweets with Year-wise Percentage Breakdown (2013-2022) Across Four Classification Approaches}
\renewcommand{\arraystretch}{1.2}
\setlength{\tabcolsep}{3pt} 
\begin{tabular}{@{}clrrr*{10}{r}@{}}
\toprule
\textbf{App.} & \textbf{L} & \textbf{Total} & \textbf{R} & \textbf{I} & \textbf{'13} & \textbf{'14} & \textbf{'15} & \textbf{'16} & \textbf{'17} & \textbf{'18} & \textbf{'19} & \textbf{'20} & \textbf{'21} & \textbf{'22} \\ 
\midrule
\multirow{4}{*}{\textbf{\#1}}  
 & EN & 4,769,850 & 47.80 & 52.20 & 3.33 & 4.01 & 5.47 & 5.24 & 5.96 & 6.32 & 4.77 & 14.88 & 24.12 & 25.91 \\
 & JA & 4,364,391 & 20.48 & 79.52 & 3.40 & 7.23 & 8.64 & 11.82 & 7.20 & 5.35 & 5.88 & 9.38 & 21.89 & 19.18 \\
 & HI & 27,258 & 35.44 & 64.56 & 0.50 & 1.22 & 1.01 & 2.49 & 2.52 & 5.16 & 3.21 & 7.75 & 26.44 & 49.70 \\
 & KO & 148,040 & 21.08 & 78.92 & 4.15 & 4.42 & 5.50 & 5.07 & 5.14 & 11.81 & 6.09 & 18.05 & 15.73 & 24.01 \\
\midrule
\multirow{4}{*}{\textbf{\#2}}  
 & EN & 4,769,850 & 50.61 & 49.39 & 3.25 & 3.93 & 5.33 & 5.12 & 5.85 & 6.33 & 4.77 & 14.91 & 24.25 & 26.26 \\
 & JA & 2,459,340 & 16.16 & 83.84 & 3.29 & 6.55 & 8.95 & 10.00 & 6.38 & 5.01 & 7.35 & 10.01 & 22.32 & 20.05 \\
 & HI & 27,258 & 35.80 & 64.20 & 0.45 & 0.51 & 0.88 & 1.21 & 1.02 & 3.82 & 2.84 & 5.70 & 26.65 & 56.92 \\
 & KO & 148,040 & 17.92 & 82.08 & 4.37 & 4.86 & 6.11 & 5.62 & 5.86 & 12.25 & 5.92 & 17.32 & 15.50 & 22.18 \\
\midrule
\multirow{4}{*}{\textbf{\#3}}  
 & EN & 4,769,850 & 48.76 & 51.24 & 3.30 & 3.99 & 5.44 & 5.22 & 5.98 & 6.28 & 4.69 & 14.88 & 24.20 & 26.02 \\
 & JA & 4,364,391 & 0.44 & 99.56 & 2.54 & 5.67 & 6.03 & 4.03 & 4.15 & 4.75 & 4.17 & 11.23 & 29.10 & 28.30 \\
 & HI & 27,258 & 5.01 & 94.99 & 0.29 & 0.07 & 1.03 & 0.51 & 0.66 & 1.98 & 1.03 & 4.18 & 21.10 & 69.15 \\
 & KO & 148,040 & 0.64 & 99.36 & 3.98 & 5.03 & 3.98 & 6.08 & 7.55 & 15.62 & 5.45 & 19.39 & 18.03 & 14.88 \\
\midrule
\multirow{4}{*}{\textbf{\#4}}  
 & EN & 4,769,850 & 49.81 & 50.19 & 3.17 & 3.89 & 5.28 & 5.08 & 5.77 & 6.34 & 4.82 & 14.96 & 24.33 & 26.36 \\
 & JA & 4,364,391 & 21.50 & 78.50 & 3.36 & 7.16 & 8.63 & 8.35 & 6.20 & 4.85 & 10.51 & 10.25 & 21.84 & 18.82 \\
 & HI & 27,258 & 31.46 & 68.54 & 0.47 & 0.43 & 0.77 & 0.75 & 0.87 & 3.43 & 2.45 & 5.56 & 27.18 & 58.09 \\
 & KO & 148,040 & 20.13 & 79.87 & 4.92 & 4.43 & 5.46 & 5.38 & 5.58 & 12.05 & 6.17 & 17.48 & 15.63 & 22.89 \\
\bottomrule
\end{tabular}
\label{tab:comprehensive_tweet_analysis}
\begin{flushleft}
\footnotesize
\textbf{Note:} App.=Approach; L=Language; R=Relevant (\%); I=Irrelevant (\%); Years shown as '13--'22 (2013--2022). All year columns show percentage of relevant tweets.
\end{flushleft}
\end{table}

\subsection{Temporal Distribution of Relevant Tweets Across Classification Approaches}
The temporal distribution of relevant tweets (Table~\ref{tab:comprehensive_tweet_analysis}) reveals a marked increase in hydrogen energy discussions from 2013–22, with post-2019 acceleration across all languages aligning with global clean energy commitments like the EU's Hydrogen Strategy (2020) \cite{EUh2Strategy} and Japan's National Hydrogen Strategy \cite{JapanH2NationalPolicy}.

\textbf{Language-specific patterns emerge clearly:} English maintains the highest relevance rates (47.80\%–50.61\%) with consistent growth peaking at 25.91\%–26.36\% in 2022. Japanese shows moderate relevance (12.36\%–20.48\%) with notable 2021 peaks (21.89\% in Approach 1). Korean exhibits similar patterns (17.92\%–21.08\% relevance) with delayed growth reaching 22.89\%–24.01\% by 2022. Hindi displays the most variable performance (5.01\%–35.80\% overall relevance) but dramatic 2022 surges (49.70\%–69.15\%), suggesting emerging market engagement despite India's complex multilingual social media landscape.

\textbf{Approach-specific insights:} Approaches 1, 2, and 4 demonstrate balanced cross-lingual performance with steady temporal growth, while Approach 3 shows significant non-English underperformance (0.44\% Japanese, 5.01\% Hindi relevance), consistent with its lower classification accuracy noted in Table~\ref{tab:approach_performance}. The post-2020 acceleration, particularly the 2021-2022 peaks, aligns with major hydrogen policy initiatives, including India's 2020 National Green Hydrogen Mission \cite{IndiaH2Strategy}.

These temporal patterns underscore hydrogen energy's evolution from niche technology to mainstream policy priority, with language-specific adoption rates reflecting regional market maturity and policy engagement levels.

\begin{figure}[t]
\centering
\includegraphics[width=0.9\textwidth]{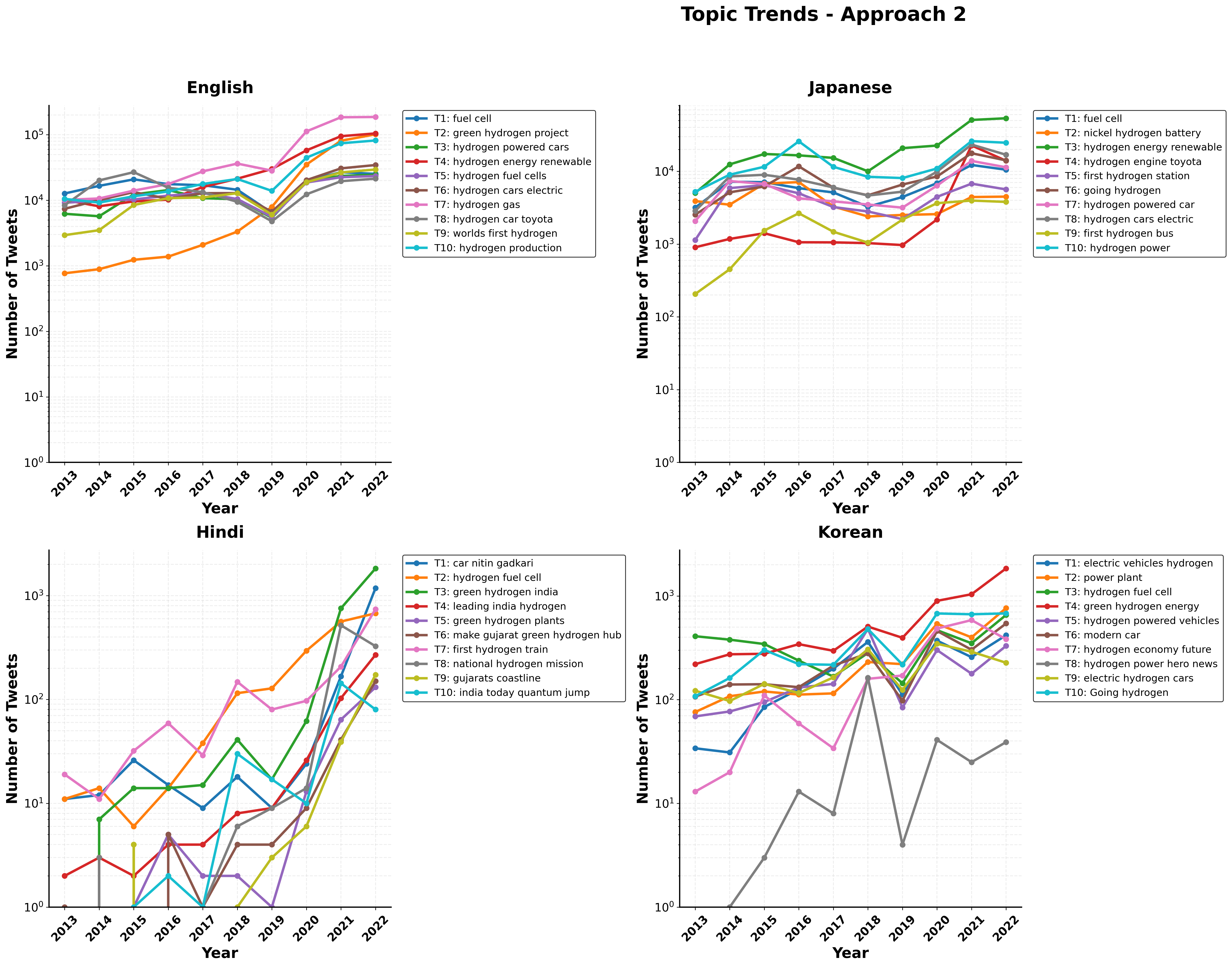}
\caption{Topic Modelling 2013-2022 on a line graph showing the trending themes across multiple languages} 
\label{fig:topic_modelling_temporal}
\end{figure}

\begin{figure}[t]
\centering
\includegraphics[width=0.9\textwidth]{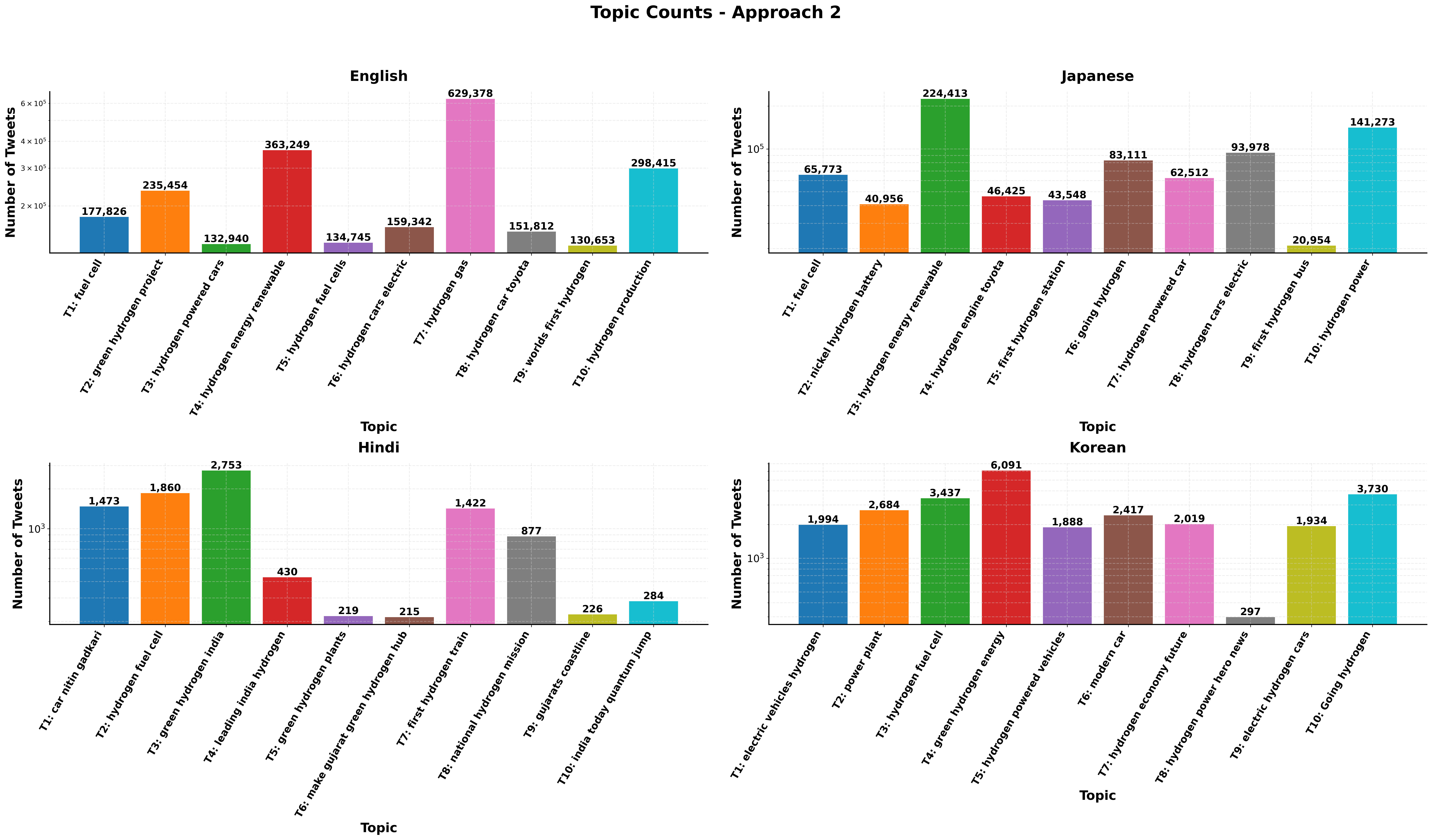}
\caption{Topic Modelling 2013-2022 on a bar graph showing the trending themes across multiple languages} 
\label{fig:topic_modelling_bar}
\end{figure}

\subsection{Topic Modeling}
\label{topic_modeling}
The topic modelling analysis of hydrogen energy discourse from 2013 to 2022 across four languages (Figures \ref{fig:topic_modelling_temporal} and \ref{fig:topic_modelling_bar}) reveals distinct thematic priorities and temporal evolution patterns under Approach 2's translation-based classification framework.

\textbf{Dominant Topics and Language-Specific Patterns:}
English discourse is dominated by ``hydrogen gas" (T7, 629,378 tweets), representing 26.0\% of total discussions, followed by ``hydrogen energy renewable" (T4, 363,249 tweets, 15.1\%) and ``hydrogen production" (T10, 298,415 tweets, 12.4\%). This distribution reflects the global emphasis on hydrogen as a primary energy carrier and its scalability in production. Japanese discussions show a more balanced distribution with ``hydrogen energy renewable" (T3, 224,413 tweets) leading at 27.3\%, followed by ``first hydrogen power" (T10, 141,273 tweets, 17.2\%), indicating Japan's focus on pioneering hydrogen technologies and renewable integration.

Hindi exhibits significantly lower absolute volumes but concentrated thematic focus, with ``green hydrogen India" (T3, 2,753 tweets, 28.2\%) and ``hydrogen fuel cell" (T2, 1,860 tweets, 19.1\%) dominating discussions, reflecting India's policy-driven approach to hydrogen adoption. ``Car nitin gadkari" (T1, 1,473 tweets, 15.09\%) likely refers to the significant event when Indian Transport Minister Nitin Gadkari arrived at Parliament in a hydrogen-powered car in March 2022, generating substantial social media attention around hydrogen vehicle adoption in India \cite{nitinGadkari}.

Korean discourse emphasises ``green hydrogen energy" (T4, 6,091 tweets, 23\%) and ``going hydrogen" (T10, 3,730 tweets, 14.1\%), suggesting infrastructure development priorities.

\textbf{Temporal Evolution Patterns}:
Temporal trends reveal language-specific growth trajectories with notable post-2019 acceleration. English topics demonstrate consistent exponential growth, with ``hydrogen gas" (T7) and ``green hydrogen project" (T2) showing dramatic increases from 2019-2022, reflecting global policy momentum. Japanese topics peaked in 2021, particularly ``hydrogen energy renewable" (T3), likely coinciding with Tokyo Olympics hydrogen initiatives and national carbon neutrality commitments, before stabilising in 2022.
Hindi shows the most dramatic late-stage growth, with several topics experiencing sharp increases in 2021-2022, particularly ``green hydrogen India" (T3) jumping from minimal presence to over 1,000 tweets, aligning with India's National Hydrogen Mission launch. Korean trends display steady growth with ``green hydrogen energy" (T4) maintaining consistent upward trajectory through 2022, suggesting sustained industrial interest.

\textbf{Strategic Implications:}
The analysis reveals three distinct regional focus areas: global scalability (English emphasis on production and gas applications), technological leadership (Japanese focus on renewable integration and pioneering applications), and policy implementation (Hindi concentration on national missions and Korean infrastructure development). These patterns suggest differentiated market opportunities, with English-speaking regions prioritising large-scale deployment, Japan leading in technology innovation, India focusing on policy-driven adoption, and Korea emphasising systematic infrastructure rollout.
The temporal surge post-2020 across all languages, with peaks coinciding with major policy announcements (EU Hydrogen Strategy \cite{EUh2Strategy}, Japan's carbon neutrality pledge \cite{greenJapan}), indicates coordinated global momentum in hydrogen energy transition, offering strategic timing insights for market entry and investment decisions.

\section{Discussion and Conclusion}
\label{conclusion}
This study explored multilingual discussions on hydrogen energy by integrating cross-lingual tweet classification with topic modeling, focusing on English, Japanese, Hindi, and Korean tweets from 2013 to 2022. We evaluated four cross-lingual relevance classification approaches, using language-specific, English-only, and multilingual models, and analysed relevant tweets to understand global clean energy conversations in multilingual digital spaces.

Our results demonstrate that model performance varies significantly across approaches and languages, with the quality of translation playing a key role. Approach 2 leveraging a single BERT model trained on English data and applied to translated tweets, achieved the highest overall performance for English and competitive results for Japanese, Korean, and Hindi. This suggests that high-quality translation to English provides a robust feature representation, outperforming Approach 1 for Japanese and Hindi, where native performance was lower (e.g., Japanese: 75.79\% ± 0.49\% accuracy, 74.88\% ± 0.56\% F1-score; Hindi: 78.08\% ± 1.10\% accuracy, 78.04\% ± 1.10\% F1-score). However, Approach 1 excelled in Korean (91.40\% ± 0.31\% accuracy, 91.35\% ± 0.31\% F1-score), indicating that language-specific models can be effective when trained and tested on high-quality native data. 

Topic modeling uncovered distinct regional discourse patterns such as English emphasised production scalability (``hydrogen gas" dominating at 26.2\%), Japanese focused on renewable integration (``hydrogen energy renewable" at 28.4\%), Hindi concentrated on policy implementation, and Korean prioritised sustainable development. Temporal analysis showed coordinated global attention on hydrogen energy after 2020, aligning with major policy initiatives, indicating market maturation, with classification accuracy directly influencing topic coherence quality.

In future, we plan to explore sentiment trends over time, aspect-based sentiment analysis for market insights, and improved cross-lingual methods that better handle native-language data instead of relying on only translations.

\section*{Acknowledgements}
\label{acknowledgments}
\subsection*{Funding}

The work has been partially supported by the Future Energy Exports CRC (www.fenex.org.au), whose activities are funded by the Australian Government's Cooperative Research Centre Program. This is a FEnEx CRC Document 2025/ 22.RP4.0139.PHD-FNX-MILE0886. The authors would like to thank Queensland University of Technology (QUT) for providing the High Performance Computing facilities used in this research and for their support.
\subsection*{Publisher}
This preprint has not undergone peer review (when applicable) or any post-submission improvements or corrections. The Version of Record of this contribution is published in Data Science and Machine Learning. AusDM 2025. Communications in Computer and Information Science, vol 2765. Springer, Singapore., and is available online at https://doi.org/10.1007/978-981-95-6786-7\_34.

\bibliography{main}
\bibliographystyle{elsarticle-num}

\end{document}